\documentclass{article}
\usepackage{iclr2018_conference,times}
\PassOptionsToPackage{sort,square,comma,numbers}{natbib}

\usepackage[utf8]{inputenc} %
\usepackage[T1]{fontenc}    %
\usepackage{hyperref}       %
\usepackage{url}            %
\usepackage{booktabs}       %
\usepackage{amsfonts}       %
\usepackage{microtype}      %

\usepackage{color}
\usepackage{amsmath}
\usepackage{caption}
\usepackage{subcaption}
\usepackage{graphicx}
\usepackage{framed}

\usepackage{tikz}
\usepackage{pgfplots}
\usepackage{pgf-umlsd}
\usepackage{ifthen}
\usetikzlibrary{shapes, snakes, shapes,positioning,calc,
  decorations.pathreplacing, shapes.misc, fit}

\usepackage{floatrow}
\usepackage{wrapfig}
\usepackage[compact]{titlesec}
\newcommand{\twoLines}[2]{\begin{tabular}{@{}c@{}}\bf #1  \\ \bf #2\end{tabular} }
\newfloatcommand{capbtabbox}{table}[][\FBwidth]

\usepackage{multirow}
\usepackage{placeins}

\DeclareMathOperator*{\argmax}{argmax}

\newcommand{\ioset}[3]{ \left\{ IO^{#1}_#2 \right\}_{#1=#3} }

\title{Leveraging Grammar and Reinforcement Learning for Neural Program Synthesis}

\author{
  Rudy Bunel\thanks{Work performed at Microsoft Research}\\
  University of Oxford\\
  \texttt{rudy@robots.ox.ac.uk} \\
  \And
  Matthew Hausknecht \\
  Microsoft Research \\
  \texttt{matthew.hausknecht@microsoft.com} \\
  \And
  Jacob Devlin$^{*}$\\
  Google \\
  \texttt{jacobdevlin@google.com} \\
  \And
  Rishabh Singh \\
  Microsoft Research \\
  \texttt{risin@microsoft.com} \\
  \And
  Pushmeet Kohli$^{*}$\\
  Deepmind \\
  \texttt{pushmeet@google.com} \\
}

\iclrfinalcopy %

\begin{document}
\maketitle

\begin{abstract}
  Program synthesis is the task of automatically generating a program consistent
  with a specification. Recent years have seen proposal of a number of neural
  approaches for program synthesis, many of which adopt a sequence generation
  paradigm similar to neural machine translation, in which sequence-to-sequence
  models are trained to maximize the likelihood of known reference programs.
  While achieving impressive results, this strategy has two key limitations.
  First, it ignores \textit{Program Aliasing}: the fact that many different
  programs may satisfy a given specification (especially with incomplete
  specifications such as a few input-output examples). By maximizing the
  likelihood of only a single reference program, it penalizes many
  \emph{semantically} correct programs, which can adversely affect the
  synthesizer performance. Second, this strategy overlooks the fact that
  programs have a strict syntax that can be efficiently checked. To address the
  first limitation, we perform reinforcement learning on top of a supervised
  model with an objective that explicitly maximizes the likelihood of generating
  semantically correct programs. For addressing the second limitation, we
  introduce a training procedure that directly maximizes the probability of
  generating syntactically correct programs that fulfill the specification. We
  show that our contributions lead to improved accuracy of the models,
  especially in cases where the training data is limited.
\end{abstract}

\section{Introduction}

The task of program synthesis is to automatically generate a program that is
consistent with a specification such as a set of input-output examples, and has
been studied since the early days of Artificial
Intelligence~\citep{waldinger1969prow}. There has been a lot of recent progress
made on \emph{neural program induction}, where novel neural architectures
inspired from computation modules such as RAM, stack, CPU, turing machines, and
GPU~\citep{graves2014neural,joulin2015inferring,nram,graves2016hybrid,npi,kaiser2015neural}
have been proposed to train these architectures in an end-to-end fashion to
mimic the behavior of the desired program. While these approaches have achieved
impressive results, they do not return explicit interpretable programs, tend not
to generalize well on inputs of arbitrary length, and require a lot of examples
and computation for learning each program. To mitigate some of these
limitations, \emph{neural program synthesis}
approaches~\citep{JohnsonHMHLZG17,NeurosymbProgsynth,RobustFill} have been
recently proposed that learn explicit programs in a Domain-specific language
(DSL) from as few as five input-output examples. These approaches, instead of
using a large number of input-output examples to learn a single program, learn a
large number of different programs, each from just a few input-output examples.
During training, the correct program is provided as reference, but at test time,
the learnt model generates the program from only the input-output examples.

\begin{figure}[t!]
  \centering
  \begin{minipage}[t!]{0.40\textwidth}
    \centering
    \begin{subfigure}{\textwidth}
    \begin{subfigure}[t]{.45\textwidth}
      \includegraphics[width=\textwidth]{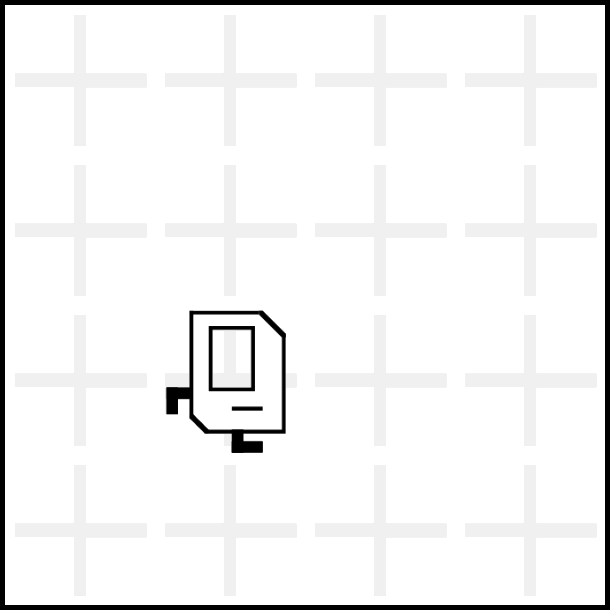}
    \end{subfigure}%
    \begin{tikzpicture}
      \useasboundingbox (0,0);
      \draw[->, ultra thick, fill=black](0,1.5) -- (.6,1.5);
    \end{tikzpicture}%
    \hfill%
    \begin{subfigure}[t]{.45\textwidth}
      \includegraphics[width=\textwidth]{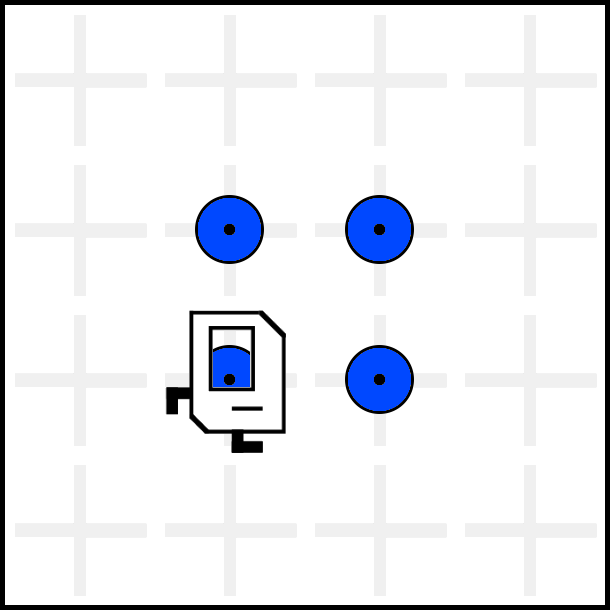}
    \end{subfigure}
    \caption{Partial Specification as IO Pair}
    \label{fig:spec}
    \end{subfigure}
  \end{minipage}%
  \hfill%
  \begin{minipage}[t!]{0.55\textwidth}
    \centering
    \scriptsize
    \begin{minipage}[t]{0.38\textwidth}
      \begin{framed}
        \centering
        {\bf \small Program A}
        \label{prog:A}
        \hrule
\begin{verbatim}
def run():
  repeat(4):
    putMarker()
    move()
    turnLeft()
\end{verbatim}
      \end{framed}
    \end{minipage}%
    \hfill %
    \begin{minipage}[t]{0.58\textwidth}
      \begin{framed}
        \centering
        {\bf \small Program B}
        \label{prog:B}
        \hrule
\begin{verbatim}
def run():
  while(noMarkersPresent):
    putMarker()
    move()
    turnLeft()
\end{verbatim}
      \end{framed}


    \end{minipage}

  \end{minipage}
  \caption{\textbf{Program Aliasing} is one difficulty of program synthesis: For
    the input-output specification given in (\ref{fig:spec}), both programs are
    \textit{semantically correct}. However, supervised training would penalize
    the prediction of Program B, if A is the ground truth.}
  \label{fig:progsynth}
\end{figure}

While neural program synthesis techniques improve over program induction
techniques in certain domains, they suffer from two key limitations. First,
these approaches use supervised learning with reference programs and suffer from
the problem of \textit{Program Aliasing}: For a small number of input-output
examples, there can be many programs that correctly transform inputs to outputs.
The problem is the discrepancy between the single supervised reference program
and the multitude of correct programs. Figure \ref{fig:progsynth} shows an
example of this: if maximizing the probability of ground truth program,
predicting Program B would be assigned a high loss even though the two programs
are semantically equivalent for the input-output example.
Maximum likelihood training forces the model to learn to predict ground truth
programs, which is different from the true objective of program synthesis:
predicting \emph{any} consistent program. To address this problem, we alter the
optimization objective: instead of maximum likelihood, we use policy gradient
reinforcement learning to directly encourage generation of \emph{any} program
that is consistent with the given examples.

The second limitation of neural program synthesis techniques based on sequence
generation paradigm~\citep{RobustFill} is that they often overlook the fact that
programs have a strict syntax, which can be checked efficiently. Similarly to
the work of~\citet{NeurosymbProgsynth}, we explore a method for leveraging the
syntax of the programming language in order to aggressively prune the
exponentially large search space of possible programs. In particular, not all
sequences of tokens are valid programs and syntactically incorrect programs can
be efficiently ignored both during training and at test time. A syntax checker
is an additional form of supervision that may not always be present. To address
this limitation, we introduce a neural architecture that retains the benefits of
aggressive syntax pruning, even without assuming access to the definition of the
grammar made in previous work~\citep{NeurosymbProgsynth}. This model
is jointly conditioned on syntactic and program correctness, and can implicitly
learn the syntax of the language while training.

We demonstrate the efficacy of our approach by developing a neural program
synthesis system for the Karel programming language~\citep{pattis1981karel}, an
educational programming language, consiting of control flow constructs such as
loops and conditionals, making it more complex than the domains tackled by
previous neural program synthesis works.

This paper makes the following key contributions:
\begin{itemize}
\item We show that Reinforcement Learning can directly optimize for generating
  any consistent program and improves performance compared to pure supervised
  learning.
\item We introduce a method for pruning the space of possible programs using a
  syntax checker and show that explicit syntax checking helps generate better
  programs.
\item In the absence of a syntax checker, we introduce a model that jointly
  learns syntax and the production of correct programs. We demonstrate this
  model improves performance in instances with limited training data.
\end{itemize}

\section{Related work}

Program synthesis is one of the fundamental problems in Artificial Intelligence.
To the best of our knowledge, it can be traced back to the work
of~\citet{waldinger1969prow} where a theorem prover was used to construct LISP
programs based on a formal specification of the input-output relation. As formal
specification is often as complex as writing the original program, many
techniques were developed to achieve the same goal with simpler partial specifications in
the form of input-output (IO)
examples~\citep{amarel1970representations,summers1977methodology}. Rule-based
synthesis approaches have recently been successful in delivering on the promise
of Programming By Example~\citep{lieberman2001your}, the most widely known
example being the FlashFill system~\citep{FlashFill} in Excel. However, such
systems are extremely complicated to extend and need significant development
time from domain experts to provide the pruning rules for efficient search.

As a result, the use of Machine Learning methods have been proposed, based on
Bayesian probabilistic models~\citep{liang2010learning} or Inductive Logic
programming~\citep{muggleton1991inductive,muggleton2014meta} to automatically
generate programs based on examples. Recently, inspired by the success of Neural
Networks in other applications such as vision~\citep{krizhevsky2012nips} or
speech recognition~\citep{graves2013speech} differentiable controllers were made
to learn the behaviour of programs by using gradient descent over differentiable
version of traditional programming concepts such as memory addressing
\citep{graves2014neural}, manipulating stacks
\citep{joulin2015inferring,grefenstette2015learning}, register machines
\citep{nram}, and data manipulation~\citep{neelakantan2015neural}. These
approaches to program induction however tend to struggle with generalization,
especially when presented with inputs of a different dimension than the one they
were trained with and require a very large amount of training data. Some
exceptions to this include Neural Programmer Interpreters~\citep{npi} and its
extensions~\citep{nplattice,cai2017making} that learn from program traces rather
than only examples. However, they still learn a different model for each program
and are computationally expensive, unlike our system that uses a single model
for learning a large number of programs.

A series of recent works aim to infer explicit program source code with the
assumption that code structure provides an inductive bias for better
generalization. In particular, explicitly modeling control flow statements such
as conditionals and loops can often lead to programs capable of generalizing,
regardless of input size~\citep{terpret,anc,diffForth}. One remaining
drawback of these approaches is the need to restart learning from scratch for
each new program. Thus they are largely unsuited for the situation of
synthesizing a new program on-the-fly from very few examples.
The latest developments use large datasets of artificially generated programs
and learn to map embeddings of IO examples to information about the programs to
generate. \citet{balog2016deepcoder} produce scores over attributes, to be used
as heuristics to speed up search-based techniques. \citet{NeurosymbProgsynth}
use their dataset to learn probability over the expansion rules of a predefined
grammar, while \citep{RobustFill} directly predict the source code of the
programs. These last two methods use supervised training to maximize the
likelihood of a single reference program, while we directly optimize for the
generation of any consistent program.

Our approach to optimize program
correctness is similar in spirit to advances in Neural Machine
Translation~\citep{googleNMT,ranzato2015sequence} that leverage
reinforcement learning to optimize directly for evaluation
metrics. Taking advantage of the fact that programs can be
syntactically checked and unit tested against the specification
examples, we show how to improve on those
REINFORCE~\citep{williams1992simple} based methods. Recently,
\citet{guu2017language} proposed a method similar to ours based on
Maximum Marginal Likelihood to generate programs based on a
description in natural language. From an application point of view,
our target domain is more complex as our DSL includes control flow
operations such as conditionals and loops.  Moreover, natural language
utterances fully describe the steps that the program needs to take,
while learning from IO examples requires planning over potentially
long executions. Their approach is more akin to inferring a formal
specification based on a natural language description, as opposed to
our generation of imperative programs.

Incorporating knowledge of the grammar of the target domain to enforce
syntactical correctness has already proven useful to model arithmetic
expressions, molecules~\citep{kusner2017grammar}, and
programs~\citep{NeurosymbProgsynth, yin17acl}. These approaches define the model over the
production rules of the grammar; we instead operate directly over the terminals
of the grammar. This allows us to learn the grammar jointly with the model in
the case where no formal grammar specification is available. Our approach is
extremely general and can be applied to the very recently proposed methods for
inferring and executing programs for visual reasoning~\citep{JohnsonHMHLZG17} that
to the best of our knowledge does not directly explictly encourage  grammar
consistency of the resulting program.

\section{Problem overview}
Before describing our proposed methods, we establish the necessary notation,
present the problem setting and describe the general paradigm of our approach.

\subsection{Program Synthesis formulation}
To avoid confusion with probabilities, we will use the letter $\lambda$ to
denote programs. $I$ and $O$ will be used to denote respectively input states
and output states and we will use the shortcut $IO$ to denote a pair of
corresponding input/output examples. A state constitutes what the programs are
going to be operating on, depending on the application domain. In FlashFill-type
applications~\citep{NeurosymbProgsynth,RobustFill}, Input and Output states would
be strings of characters, while in our Karel environment, states are grids
describing the presence of objects. If we were to apply our method to actual
programming languages, states would represent the content of the machine's
registers and the memory.

At training time, we assume to have access to $N$ training samples, each
training sample consisting of a set of $K$ Input/Output states and a program
implementing the mapping correctly:
\begin{equation}
    \mathcal{D} = \Bigg\{\ \Big( \ioset{k}{i}{1..K}, \ \lambda_i \Big)\ \Bigg\}_{i=1..N}  \text{such that: } \quad  \lambda_i(I^k_i) = O^k_i \quad \forall i \in 1 .. N, \quad \forall k \in 1 .. K
\end{equation}
where $\lambda_i(I^k_i)$ denotes the resulting state of applying the program
$\lambda_i$ to the input state $I^k_i$. Our goal is to learn a synthesizer $\sigma$
that, given a set of input/output examples produces a program:
\begin{equation}
    \sigma : \quad  \left\{ IO^k \right\}_{k=1..K} \quad \longrightarrow \quad \hat{\lambda}
\end{equation}
We evaluate the programs on a set of test cases for which we have both
specification examples and held-out examples:
\begin{equation}
    \mathcal{D}_{\text{test}}= \Bigg\{\ \Big( \ioset{k_\text{spec}}{j}{1..K} , \quad  \ioset{k_\text{test}}{j}{K+1..K^\prime}  \Big) \ \Bigg\}_{j=1..N_\text{test}}
\end{equation}
At test time, we evaluate the performance of our learned synthesizer by
generating, for each sample in the test set, a program $\hat{\lambda_j}$. The
metric we care about is \emph{Generalization}:
{\small
\begin{equation}
  \label{eq:generalization}
  \Big\{\ j \in \{1..N_\text{test}\}\ \text{such that } \  \hat{\lambda_j}(I^k_j) == O^k_j \quad \forall k \in \left\{1..K^\prime\right\} \Big\} \text{ where } \hat{\lambda_j} = \sigma\left( \ioset{k_\text{spec}}{j}{1..K} \right).
\end{equation}
}

\subsection{Neural Program Synthesis Architecture}
Similar to \citet{RobustFill} we use a sequential LSTM-based~\citep{lstm} language model, conditioned on an
embedding of the input-output pairs. Each pair is encoded independently by a convolutional neural network (CNN)
to generate a joint embedding. A succinct description of the architecture can be
found in section \ref{subsec:exp_domain} and the exact dimensions are available in
the supplementary materials.

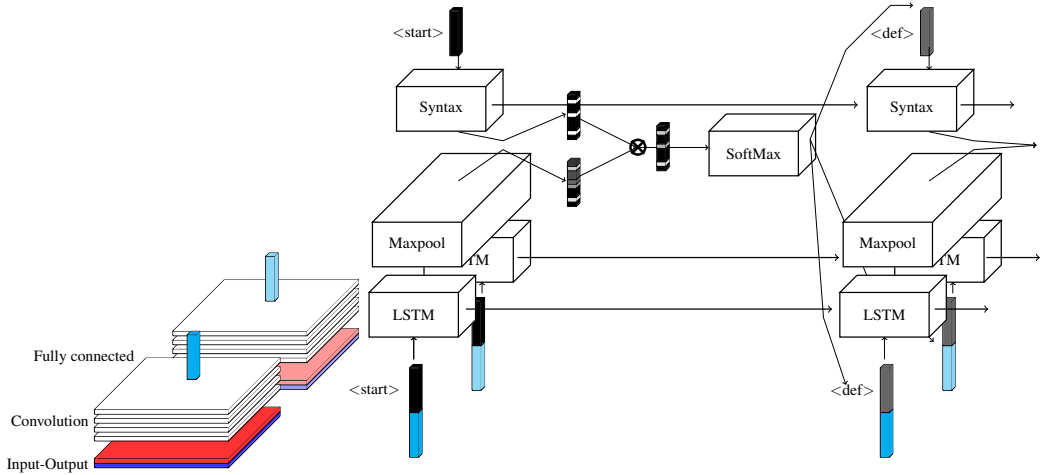
\begin{figure}[t]
  {  \noindent\resizebox{\textwidth}{!}{
      \begin{tikzpicture}

  \newcommand{\cube}[7]{
    \def\L{-#1} 
    \def\H{#2} 
    \def\D{#3} 
    \def\X{#4} 
    \def\Y{#5} 
    \def\Z{#6} 
    \def\content{#7} 

    \filldraw[color=white] (\X+\L,\Y+\H,\Z)--(\X,\Y+\H,\Z)--(\X,\Y+\H,\Z+\D)--(\X+\L,\Y+\H,\Z+\D)--(\X+\L,\Y+\H,\Z);
    \filldraw[color=white] (\X+\L,\Y+\H,\Z+\D)--(\X+\L,\Y,\Z+\D)--(\X,\Y,\Z+\D)--(\X,\Y+\H,\Z+\D)--(\X+\L,\Y+\H,\Z+\D);
    \filldraw[color=white] (\X,\Y,\Z+\D)--(\X,\Y,\Z)--(\X,\Y+\H,\Z)--(\X,\Y+\H,\Z+\D)--(\X,\Y,\Z+\D);

    \draw[thick](\X+\L,\Y+\H,\Z)--(\X,\Y+\H,\Z)--(\X,\Y+\H,\Z+\D)--(\X+\L,\Y+\H,\Z+\D)--(\X+\L,\Y+\H,\Z);
    \draw[thick](\X+\L,\Y+\H,\Z+\D)--(\X+\L,\Y,\Z+\D)--(\X,\Y,\Z+\D)--(\X,\Y+\H,\Z+\D);
    \draw[thick](\X,\Y,\Z+\D)--(\X,\Y,\Z)--(\X,\Y+\H,\Z);
    \draw(\X+0.5*\L,\Y+0.5*\H,\Z+\D) node{\content};
  }

  \newcommand{\networkLayer}[6]{
    \def\a{#1} 
    \def\b{0.02}
    \def\c{#2} 
    \def\t{#3} 
    \ifthenelse {\equal{#6} {}} {\def\y{0}} {\def\y{#6}} 

    \draw[line width=0.25mm](\c+\t,\y,0) -- (\c+\t,\a+\y,0) -- (\t,\a+\y,0);                                                      
    \draw[line width=0.25mm](\t,\y,\a) -- (\c+\t,\y,\a) node[midway,left] {#5} -- (\c+\t,\a+\y,\a) -- (\t,\a+\y,\a) -- (\t,\y,\a); 
    \draw[line width=0.25mm](\c+\t,\y,0) -- (\c+\t,\y,\a);
    \draw[line width=0.25mm](\c+\t,\a+\y,0) -- (\c+\t,\a+\y,\a);
    \draw[line width=0.25mm](\t,\a+\y,0) -- (\t,\a+\y,\a);

    \filldraw[#4] (\t+\b,\b+\y,\a) -- (\c+\t-\b,\b+\y,\a) -- (\c+\t-\b,\a-\b+\y,\a) -- (\t+\b,\a-\b+\y,\a) -- (\t+\b,\b+\y,\a); 
    \filldraw[#4] (\t+\b,\a+\y,\a-\b) -- (\c+\t-\b,\a+\y,\a-\b) -- (\c+\t-\b,\a+\y,\b) -- (\t+\b,\a+\y,\b);

    \ifthenelse {\equal{#4} {}}
    {} 
    {\filldraw[#4] (\c+\t,\b+\y,\a-\b) -- (\c+\t,\b+\y,\b) -- (\c+\t,\a-\b+\y,\b) -- (\c+\t,\a-\b+\y,\a-\b);} 
  }
  \begin{scope}[rotate=90, xshift=50, yshift=60, yscale=-1]
    \networkLayer{3.0}{0.1}{-0.4}{color=blue!40}{}{}
    \networkLayer{3.0}{0.1}{-0.3}{color=red!40}{}{}

    \networkLayer{3.0}{0.1}{0.2}{color=white}{}{}    
    \networkLayer{3.0}{0.1}{0.4}{color=white}{}{}        
    \networkLayer{3.0}{0.1}{0.6}{color=white}{}{}        
    \networkLayer{3.0}{0.1}{0.8}{color=white}{}{}        
    \networkLayer{0.2}{1}{0.5}{color=cyan!40}{}{1}        
  \end{scope}

  \begin{scope}[rotate=90, yshift=110, yscale=-1]
    \networkLayer{3.0}{0.1}{-0.4}{color=blue!80}{Input-Output}{}
    \networkLayer{3.0}{0.1}{-0.3}{color=red!80}{}{}

    \networkLayer{3.0}{0.1}{0.2}{color=white}{}{}    
    \networkLayer{3.0}{0.1}{0.4}{color=white}{}{}        
    \networkLayer{3.0}{0.1}{0.6}{color=white}{Convolution}{}        
    \networkLayer{3.0}{0.1}{0.8}{color=white}{}{}        
    \networkLayer{0.2}{1}{0.5}{color=cyan}{Fully connected$\qquad\quad$}{1}        
  \end{scope}

  \begin{scope}[xshift=100, yshift=50]
    \cube{2}{1}{1}{2}{2}{2}{LSTM}
    \begin{scope}[rotate=90, yscale=-1]
      \networkLayer{0.2}{1}{-1.5}{color=cyan!40}{}{0}
      \networkLayer{0.2}{1}{-0.5}{color=black}{}{0}
    \end{scope}
    \draw[->](1.3,1.75,3) -- (1.3,2,3);

    \cube{2}{1}{1}{0}{0}{0}{LSTM}
    \begin{scope}[rotate=90, yscale=-1]
      \networkLayer{0.2}{1}{-3}{color=cyan}{}{-1.4}
      \networkLayer{0.2}{1}{-2}{color=black}{$<$start$>\ $}{-1.4}
    \end{scope}
    \draw[->](-1,-.5,1) -- (-1,0,1);

    \cube{2}{1}{4}{2}{3.5}{2}{Maxpool}
    \draw[->](-1,-.5,1) -- (-1,0,1);

    \begin{scope}[rotate=90, yscale=-1]
      \networkLayer{0.2}{1}{6}{color=black}{$<$start$>$}{-0.5}
    \end{scope}
    \cube{2}{1}{1}{1}{5}{1}{Syntax}
    \draw[->](0, 6.5, 1) -- (0,6,1);
    \draw[->](0,4.6,1) --  (1, 4.4, 1) -- (2.4,   5, 1);
    \draw[->](0,3.5,1) --  (1, 4.2, 1) -- (2.4, 3.5, 1);

    \draw[->](2.5, 5,  1) -- (4, 4.25, 1);
    \draw[->](2.5, 3.5,1) -- (4, 4.25, 1);

    \draw[->](4, 4.25, 1)--(5.6, 4.25, 1);

    \begin{scope}[rotate=90, yscale=-1, yshift=75, xshift=-25]
      \networkLayer{0.2}{.1}{5}{color=black}{}{-0.5}
      \networkLayer{0.2}{.1}{5.1}{color=white}{}{-0.5}
      \networkLayer{0.2}{.3}{5.2}{color=black}{}{-0.5}
      \networkLayer{0.2}{.1}{5.5}{color=white}{}{-0.5}
      \networkLayer{0.2}{.2}{5.6}{color=black}{}{-0.5}
      \networkLayer{0.2}{.1}{5.8}{color=white}{}{-0.5}
      \networkLayer{0.2}{.1}{5.9}{color=black}{}{-0.5}

      \networkLayer{0.2}{.1}{3.5}{color=black}{}{-0.5}
      \networkLayer{0.2}{.1}{3.6}{color=black!10}{}{-0.5}
      \networkLayer{0.2}{.2}{3.7}{color=black}{}{-0.5}
      \networkLayer{0.2}{.1}{3.9}{color=black!40}{}{-0.5}
      \networkLayer{0.2}{.1}{4}{color=black!60}{}{-0.5}
      \networkLayer{0.2}{.2}{4.1}{color=black!70}{}{-0.5}
      \networkLayer{0.2}{.1}{4.3}{color=black!20}{}{-0.5}
      \networkLayer{0.2}{.1}{4.4}{color=black!80}{}{-0.5}

      \node[draw, ultra thick, cross out] at (4.75, 1, 0) {};
      \node[draw, ultra thick, circle] at (4.75, 1, 0) {};

      \networkLayer{0.2}{.1}{4.3}{color=black}{}{1.5}
      \networkLayer{0.2}{.1}{4.4}{color=black!10}{}{1.5}
      \networkLayer{0.2}{.3}{4.5}{color=black}{}{1.5}
      \networkLayer{0.2}{.1}{4.8}{color=black!30}{}{1.5}
      \networkLayer{0.2}{.2}{4.9}{color=black}{}{1.5}
      \networkLayer{0.2}{.1}{5.1}{color=black!20}{}{1.5}
      \networkLayer{0.2}{.1}{5.2}{color=black}{}{1.5}
    \end{scope}

    \cube{2}{1}{1}{8}{4}{1}{SoftMax}

  \end{scope}

  \draw[->](11.2,6,.5) -- (12, 8, .5) -- (13, 9, .5) -- (13.5, 9, .5);
  \draw[->](11.2,6,.5) -- (11.5, 2, .5) -- (12, .5, .5) ;
  \draw[->](11.2,6,.5) -- (11.5, 2, -2) -- (13, .5, -2) ;

  \draw[thick,->](3.3, 2, 0) -- (11.5, 2, 0);
  \draw[thick,->](3.3, 2, -3) -- (10.5, 2, -3);
  \draw[thick,->](3.3, 6, -1.5) -- (11.5, 6, -1.5);

  \begin{scope}[xshift=400, yshift=50]
    \cube{2}{1}{1}{2}{2}{2}{LSTM}
    \begin{scope}[rotate=90, yscale=-1]
      \networkLayer{0.2}{1}{-1.5}{color=cyan!40}{}{0}
      \networkLayer{0.2}{1}{-0.5}{color=black!60}{}{0}
    \end{scope}
    \draw[->](1.3,1.75,3) -- (1.3,2,3);

    \cube{2}{1}{1}{0}{0}{0}{LSTM}
    \begin{scope}[rotate=90, yscale=-1]
      \networkLayer{0.2}{1}{-3}{color=cyan}{}{-1.4}
      \networkLayer{0.2}{1}{-2}{color=black!60}{$<$def$>$}{-1.4}
    \end{scope}
    \draw[->](-1,-.5,1) -- (-1,0,1);

    \cube{2}{1}{4}{2}{3.5}{2}{Maxpool}

    \begin{scope}[rotate=90, yscale=-1]
      \networkLayer{0.2}{1}{6}{color=black!60}{$<$def$>$}{-0.5}
    \end{scope}
    \cube{2}{1}{1}{1}{5}{1}{Syntax}
    \draw[->](0, 6.5, 1) -- (0,6,1);
    \draw[->](0,4.6,1) --  (1, 4.4, 1) -- (2.4, 4.3, 1);
    \draw[->](0,3.5,1) --  (1, 4.2, 1) -- (2.4, 4.3, 1);
  \end{scope}

  \draw[thick,->](13.8, 2, 0) -- (15, 2, 0);
  \draw[thick,->](13.8, 2, -3) -- (15, 2, -3);
  \draw[thick,->](13.8, 6, -1.5) -- (15, 6, -1.5);

\end{tikzpicture}
    }}
  \caption{Architecture of our model. Each pair of Input-Output is embedded
    jointly by a CNN. One decoder LSTM is run for each example, getting fed in a
    concatenation of the previous token and the IO pair embedding (constant
    across timestep). Results of all the decoders are maxpooled and the
    prediction is modulated by the mask generated by the syntax model. The
    probability over the next token is then obtained by a Softmax transformation.}
  \label{fig:model_arch}
\end{figure}

Each program is represented by a sequence of tokens $\lambda = [s_1, s_2, ...,
s_L]$ where each token comes from an alphabet $\Sigma$. We model the program one
token at a time using an LSTM. At each timestep, the input consists of the
concatenation of the embedding of the IO pair and of the last predicted token.
One such decoder LSTM is run for each of the IO pairs, all using the same
weights. The probability of the next token is defined as the Softmax of a linear
layer over the max-pooled hidden state of all the decoder LSTMs. A schema
representing this architecture can be seen in Figure~\ref{fig:model_arch}.

The form of the model that we are learning is:
{\small
\begin{equation}
  p_{\theta}(\lambda_i \ | \ \ioset{k}{i}{1..K}) = \prod_{t=1}^{L_i}\ p_\theta(s_t \ | \ s_1, ...,s_{t-1}, \ioset{k}{i}{1..K})
\end{equation}
}
At test time, the most likely programs are obtained by running a beam search. One
of the advantages of program synthesis is the ability to execute hypothesized programs. Through execution, we remove syntactically incorrect programs
and programs that are not consistent with the observed examples. Among the remaining programs, we
return the most likely according to the model.

\section{Objective Functions}

\subsection{Maximum Likelihood optimization}
To estimate the parameters $\theta$ of our model, the default solution is to
perform supervised training, framing the problem as Maximum Likelihood
estimation. \citet{RobustFill} follow this approach and use stochastic gradient
descent to solve:
\begin{equation}
  \label{eq:mle-obj}
  \theta^{\star} = \argmax_{\theta} \prod_{i=1..N} p_{\theta}(\lambda_i\ | \ \ioset{i}{i}{1..K}) = \argmax_{\theta} \sum_{i=1..N} \log \left( p_{\theta}( \lambda_i \ | \  \ioset{k}{i}{1..K})  \right)
\end{equation}

However, this training objective exhibits several drawbacks. First, at
training time, the model is only exposed to the training data distribution,
while at test time, it is fed back the token from its own previous predictions.
This discrepancy in distribution of the inputs is well known in Natural Language
Processing under the name of \textit{exposure bias}~\citep{ranzato2015sequence}.

Moreover, this loss does not represent the true objective of program synthesis.
In practice, any equivalent program should be as valid a prediction as the
reference one. This property, that we call \textit{program aliasing}, is not
taken into account by the MLE training.
Ideally, we would like the model to learn to reason about the necessary steps
needed to map the input to the output. As a result, the loss shouldn't penalize
correct programs, even if they do not correspond to the ground truth.

\subsection{Optimizing Expected Correctness}
\label{subsec:expected-correctness}
The first modification that we propose is to change the target objective to
bring it more in line with the goal of program synthesis. We replace the
optimization problem of (\ref{eq:mle-obj}) by
\begin{equation}
  \label{eq:RL}
  \theta^{\star} = \argmax_{\theta} \mathcal{L}_{R}(\theta), \qquad \text{ where } \mathcal{L}_{R}(\theta) = \sum_{i..N} \left( \sum_{\lambda} p_{\theta}(\lambda \ | \ioset{k}{i}{1..K})\ R_i(\lambda)\right),
\end{equation}
where $R_i(\lambda)$ is a reward function designed to encode the quality of
the sampled programs. Note that this formulation is extremely generic and would
allow to represent a wide range of different objective functions. If we assume
that we have access to a simulator to run our programs, we can design $R_i$ so
as to optimize for generalization on held-out examples, preventing the model to
overfit on its inputs. Additional property such as program conciseness, or
runtime efficiency could also be encoded into the reward.

However, this expressiveness comes at a cost: the inner sum in (\ref{eq:RL}) is
over all possible programs and therefore is not tractable to compute. The
standard method consists of approximating the objective by defining a Monte
Carlo estimate of the expected reward, using $S$ samples from the model. To
perform optimization, an estimator of the gradient of the expected reward is
built based on the REINFORCE trick~\citep{williams1992simple}.
\begin{equation}
  \begin{split}
  \label{eq:Rl-sample}
  \mathcal{L}_{R}(\theta) &\approx \sum_{i=1..N}\sum_{r=1}^{S} \frac{1}{S}\ R_i(\lambda_r), \qquad \text{ where } \lambda_r \sim p_\theta(\  \cdot \  |  \ioset{k}{i}{1..K})\\
  \nabla_{\theta} \mathcal{L}_R(\theta) &\approx \sum_{i=1..N} \sum_{r=1}^{S} \frac{1}{S} \ R_i(\lambda_r) \nabla_{\theta}\log\left(  p_\theta(\  \lambda_r \  |  \ioset{k}{i}{1..K}) \right)
   \end{split}
\end{equation}

However, given that we sample from a unique model, there is a high chance that
we will sample the same programs repeatedly when estimating the gradient. This
is especially true when the model has been pre-trained in a supervised manner. A
different approach is to approximate the distribution of the learned
distribution by another one with a smaller support.

\begin{figure}[t]
	\noindent\resizebox{\textwidth}{!}{
    \begin{tikzpicture}

  \def\cellsize{2pt}
  \newcommand{\cell}[3]{
    \def\name{#1}
    \def\color{#2}
    \def\prevcell{#3}

    \ifthenelse {\equal{#3} {}} {
      \node[draw, rectangle, fill=\color, minimum width=\cellsize, minimum height=\cellsize] (\name){};
    }{
      \node[draw, rectangle, fill=\color, minimum width=\cellsize, minimum height=\cellsize, right = 0 of \prevcell.east] (\name){};
    }
  }

  \begin{scope}[]
    \cell{c010}{blue}{}
    \cell{c011}{red}{c010}
    \cell{c012}{blue}{c011}
    \cell{c013}{red}{c012}

    \node[left = 0 of c010.west] (c01label) {C$_{1, t}$};
    \node[right = 0 of c013.east] (c01score) {0.5};
  \end{scope}

  \begin{scope}[yshift=-10]
    \cell{c020}{blue}{}
    \cell{c021}{blue}{c020}
    \cell{c022}{red}{c021}
    \cell{c023}{red}{c022}

    \node[left = 0 of c020.west] (c02label) {C$_{2, t}$};
    \node[right = 0 of c023.east] (c02score) {0.2};
  \end{scope}

  \begin{scope}[yshift=-20]
    \cell{c030}{red}{}
    \cell{c031}{blue}{c030}
    \cell{c032}{blue}{c031}
    \cell{c033}{red}{c032}

    \node[left = 0 of c030.west] {C$_{3, t}$};
    \node[right = 0 of c033.east] (c03score) {0.1};
  \end{scope}

  \node[ultra thick, left = of c02label.west]{$\cdots$};

  \node[draw, dotted, fit= (c01label) (c033)] (ctbox){};
  \node[below = 0 of ctbox.south] {Step $t$ candidates};

  \begin{scope}[xshift=60, yshift=30]
    \cell{c110}{blue}{}
    \cell{c111}{red}{c110}
    \cell{c112}{blue}{c111}
    \cell{c113}{red}{c112}
    \cell{c114}{blue}{c113}
    \node[right = 0 of c114.east] (c11score) {0.08};
    \node[circle, ultra thick, draw=olive, right = 0 of c11score, inner sep=2] (pick1) {};
  \end{scope}
  \begin{scope}[xshift=60, yshift=22]
    \cell{c120}{blue}{}
    \cell{c121}{red}{c120}
    \cell{c122}{blue}{c121}
    \cell{c123}{red}{c122}
    \cell{c124}{red}{c123}
    \node[right = 0 of c124.east] (c12score) {0.35};
    \node[circle, ultra thick, draw=olive, right = 0 of c12score, inner sep=2] (pick2) {};
  \end{scope}
  \begin{scope}[xshift=60, yshift=14]
    \cell{c130}{blue}{}
    \cell{c131}{red}{c130}
    \cell{c132}{blue}{c131}
    \cell{c133}{red}{c132}
    \cell{c134}{green}{c133}
    \node[right = 0 of c134.east] (c13score) {0.07};
    \node[cross out, ultra thick, draw=olive, right = 0 of c13score, inner sep=2] {};
  \end{scope}

  \begin{scope}[xshift=60, yshift=-2]
    \cell{c140}{blue}{}
    \cell{c141}{blue}{c140}
    \cell{c142}{red}{c141}
    \cell{c143}{red}{c142}
    \cell{c144}{blue}{c143}
    \node[right = 0 of c144.east] (c14score) {0};
  \end{scope}
  \begin{scope}[xshift=60, yshift=-10]
    \cell{c150}{blue}{}
    \cell{c151}{blue}{c150}
    \cell{c152}{red}{c151}
    \cell{c153}{red}{c152}
    \cell{c154}{red}{c153}
    \node[right = 0 of c154.east] (c15score) {0.10};
    \node[circle, ultra thick, draw=olive, right = 0 of c15score, inner sep=2] (pick3) {};
  \end{scope}
  \begin{scope}[xshift=60, yshift=-18]
    \cell{c160}{blue}{}
    \cell{c161}{blue}{c160}
    \cell{c162}{red}{c161}
    \cell{c163}{red}{c162}
    \cell{c164}{green}{c163}
    \node[right = 0 of c164.east] (c16score) {0.10};
    \node[cross out, ultra thick, draw=olive, right = 0 of c16score, inner sep=2] {};
  \end{scope}

  \begin{scope}[xshift=60, yshift=-34]
    \cell{c170}{red}{}
    \cell{c171}{blue}{c170}
    \cell{c172}{blue}{c171}
    \cell{c173}{red}{c172}
    \cell{c174}{blue}{c173}
    \node[right = 0 of c174.east] (c17score) {0.05};
  \end{scope}
  \begin{scope}[xshift=60, yshift=-42]
    \cell{c180}{red}{}
    \cell{c181}{blue}{c180}
    \cell{c182}{blue}{c181}
    \cell{c183}{red}{c182}
    \cell{c184}{red}{c183}
    \node[right = 0 of c184.east] (c18score) {0.02};
  \end{scope}
  \begin{scope}[xshift=60, yshift=-50]
    \cell{c190}{red}{}
    \cell{c191}{blue}{c190}
    \cell{c192}{blue}{c191}
    \cell{c193}{red}{c192}
    \cell{c194}{green}{c193}
    \node[right = 0 of c194.east] (c19score) {0.03};
    \node[cross out, ultra thick, draw=olive, right = 0 of c19score, inner sep=2] {};
  \end{scope}

  \draw[->] (c01score.east) -- (c120.west);
  \draw[->] (c02score.east) -- (c150.west);
  \draw[->] (c03score.east) -- (c180.west);

   \begin{scope}[xshift=180]
    \cell{c210}{blue}{}
    \cell{c211}{red}{c210}
    \cell{c212}{blue}{c211}
    \cell{c213}{red}{c212}
    \cell{c214}{blue}{c213}

    \node[left = 0 of c210.west] (c21label) {C$_{1, t+1}$};
    \node[right = 0 of c214.east] (c21score) {0.08};
  \end{scope}

  \begin{scope}[xshift=180, yshift=-10]
    \cell{c220}{blue}{}
    \cell{c221}{red}{c220}
    \cell{c222}{blue}{c221}
    \cell{c223}{red}{c222}
    \cell{c224}{red}{c223}

    \node[left = 0 of c220.west] (c22label) {C$_{2, t+1}$};
    \node[right = 0 of c224.east] (c22score) {0.35};
  \end{scope}

  \begin{scope}[xshift=180, yshift=-20]
    \cell{c230}{blue}{}
    \cell{c231}{blue}{c230}
    \cell{c232}{red}{c231}
    \cell{c233}{red}{c232}
    \cell{c234}{red}{c233}

    \node[left = 0 of c230.west] (c23label) {C$_{3, t+1}$};
    \node[right = 0 of c234.east] (c23score) {0.10};
  \end{scope}

  \draw[->] (pick1.east) -- (c21label.west);
  \draw[->] (pick2.east) -- (c22label.west);
  \draw[->] (pick3.east) -- (c23label.west);

  \node[draw, dotted, fit= (c21label) (c234.south east)] (ctp1box) {};
  \node[below = 0 of ctp1box.south] {Step $t+1$ candidates};

  \node[above right = 0.5 and 0.5 of c21score] (dummyc21) {};
  \draw[->] (c21score) -- (dummyc21);
  \node[right = 0.5 of c22score] (dummyc22) {};
  \draw[->] (c22score) -- (dummyc22);
  \node[below right = 0.5 and 0.5 of c23score] (dummyc23) {};
  \draw[->] (c23score) -- (dummyc23);

  \node[right = 0 of dummyc22](dots) {$\cdots\ \cdots$};

\begin{scope}[xshift=320]
    \cell{c310}{blue}{}
    \cell{c311}{red}{c310}
    \cell{c312}{blue}{c311}
    \cell{c313}{red}{c312}
    \cell{c314}{red}{c313}
    \cell{c315}{red}{c314}
    \cell{c316}{green}{c315}

    \node[right = 1 of c316.east] (c31score) {0.20};
  \end{scope}

  \begin{scope}[xshift=320, yshift=-10]
    \cell{c320}{blue}{}
    \cell{c321}{red}{c320}
    \cell{c322}{blue}{c321}
    \cell{c323}{red}{c322}
    \cell{c324}{red}{c323}
    \cell{c325}{blue}{c324}
    \cell{c326}{green}{c325}

    \node[right = 1 of c326.east] (c32score) {0.10};
  \end{scope}

  \begin{scope}[xshift=320, yshift=-20]
    \cell{c330}{blue}{}
    \cell{c331}{blue}{c330}
    \cell{c332}{red}{c331}
    \cell{c333}{red}{c332}
    \cell{c334}{green}{c333}

    \node[below = -0.1 of c32score] (c33score) {0.10};
  \end{scope}

  \node[draw, dotted, fit= (c310) (c316) (c330)](finalbox){};
  \node[below = 0 of finalbox.south] (finalboxlabel) {BS($p_\theta$, 3)};

  \node[draw, dotted, fit= (c31score) (c33score)](pthetabox){};
  \node[below = 0 of pthetabox.south] (pthetaboxlabel) {$p_\theta\left( \cdot | \{IO\} \right)$};

  \node[right = 1.5 of c31score] (rescscore1) {0.50};
  \node[right = 1.5 of c32score] (rescscore2) {0.25};
  \node[right = 1.5 of c33score] (rescscore3) {0.25};

  \draw[->] (c31score) -- (rescscore1);
  \draw[->] (c32score) -- (rescscore2);
  \draw[->] (c33score) -- (rescscore3);

  \node[draw, dotted, fit= (rescscore1) (rescscore3)](qthetabox){};
  \node[below = 0 of qthetabox.south] (qthetaboxlabel) {$q_\theta\left( \cdot | \{IO\} \right)$};
\end{tikzpicture}
		}
    \caption{\label{fig:beamsearch}Approximation using a beamsearch. All
      possibles next tokens are tried for each candidates, the $S$ (here 3) most
      likely according to $p_\theta$ are kept. When an End-Of-Sequence token
      (green) is reached, the candidate is held out. At the end, the most likely
      complete sequences are used to construct an approximate distribution,
      through rescaling.}
\end{figure}
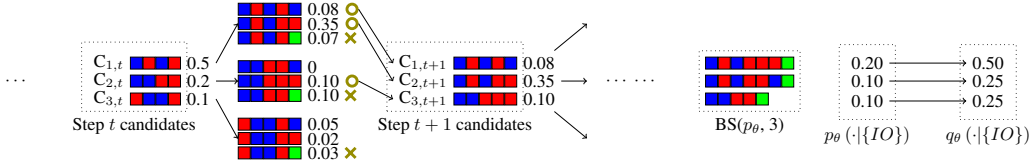

To obtain this smaller distribution, one possible solution is to employ the $S$
most likely samples as returned by a Beam Search. We generate the embedding of
the IO grids and perform decoding, keeping at each step the $S$ most likely
candidates prefixes based on the probability $p_{\theta}$ given by the model. At
step $t$, we evaluate $p_{\theta}(s_1 \dots s_t, \ioset{k}{i}{1..K})$ for all
the possible next token $s_t$ and all the candidates $(s_1 \dots s_{t-1})$
previously obtained. The $S$ most likely sequences will be the candidates at the
$(t+1)$ step. Figure \ref{fig:beamsearch} represents this process.

Based on the final samples obtained, we define a probability distribution to use
as an approximation of $p_{\theta}$ in \eqref{eq:RL}. As opposed to
\eqref{eq:Rl-sample}, this approximation introduces a bias. It however has the
advantage of aligning the training procedure more closely with the testing
procedure where only likely samples are going to be decoded. Formally, this
corresponds to performing the following approximation of the objective function,
($\text{BS}(p_\theta, S)$ being the $S$ samples returned by a beam search with
beam size $S$):
\begin{equation}
  \label{eq:rl-bs}
  \begin{split}
    & \mathcal{L}_R(\theta) \approx \sum_{i..N} \left( \sum_{\lambda} q_{\theta}(\lambda \ | \ioset{k}{i}{1..K})\ R_i(\lambda)\right)\\
    & \text{where } \qquad q_{\theta}(\lambda_r \ | \ioset{k}{i}{1..K}) = \begin{cases}
      \frac{p_{\theta}(\lambda_r \ | \ioset{k}{i}{1..K})}{\sum_{\lambda_r \in \text{BS}(p_{\theta}, S)} p_{\theta}(\lambda_r \ | \ioset{k}{i}{1..K})} &\qquad \text{ if } \lambda_r \in \text{BS}(p_{\theta}, S)\\
      0 &\qquad \text{otherwise}.
      \end{cases}
  \end{split}
\end{equation}
With this approximation, the support of the distribution $q_{\theta}$ is much
smaller as it contains only the $S$ elements returned by the beam search. As a
result, the sum become tractable and no estimator are needed. We can simply
differentiate this new objective function to obtain the gradients of the loss
with regards to $p_\theta$ and use the chain-rule to subsequently obtain
gradient with regards to $\theta$ necessary for the optimization.
Note that if $S$ is large enough to cover all possible programs, we recover the
objective of \eqref{eq:RL}.

Based on this more tractable distribution, we can define more complex objective
functions. In program synthesis, we have the possibility to prune out several
predictions by using the specification. Therefore, we can choose to go beyond
optimizing the expected reward when sampling a single program and optimize the
expected reward when sampling a bag of $C$ programs and keeping the best one.
This results in a new objective function: {\small
  \begin{equation}
    \label{eq:rl-div-gen}
    \theta^{\star} = \argmax_{\theta} \sum_{i=1..N} \left( \sum_{ \{\lambda_{1},..,\lambda_{C}\} \in {\text{BS}(p_{\theta, s})^C}} \left[ \max_{j \in {1..C}} R_i(\lambda_j)\right] \left( \prod_{r \in {1..C}} q_{\theta}\left(\lambda_r \ | \ioset{k}{i}{1..K}\right)  \right)  \right),
  \end{equation}}
where $q_{\theta}$ is defined as previously. We argue that by optimizing this
objective function, the model gets the capability of ``hedging its bets'' and
assigning probability mass to several candidates programs, resulting in a higher
diversity of outputs.

In the special case where the reward function only takes values in $\{0, 1\}$,
as it is when we are using correctness as a reward, this can be more easily computed as:
{\small\begin{equation}
  \label{eq:rl-div-zeroone}
    \theta^{\star} = \argmax_{\theta} \sum_{i=1..N} \left(1  - \left(\sum_{\lambda_r \in \text{BS}(p_{\theta}, S)} [R_i(\lambda_r) == 0]\  q_{\theta}(\lambda_r \ | \ioset{k}{i}{1..K}) \right)^C \right)
  \end{equation}}
The derivation leading to this formulation as well as a description on how to
efficiently compute the more general loss\eqref{eq:rl-div-gen} can be found in
appendix \ref{app:div_details}.
Note that although this formulation brings the training objective function
closer to the testing procedure, it is still not ideal. It indeed makes the
assumption that if we have a correct program in our bag of samples, we can
identify it, ignoring the fact that it is possible to have some incorrect program
consistent with the IO pairs partial specification (and therefore not prunable).
In addition, this represents a probability where the $C$ programs are sampled
independently, which in practice we wouldn't do.

\section{Model}
\subsection{Conditioning on the syntax}
\label{subsec:handwritten}
One aspect of the program synthesis task is that syntactically incorrect
programs can be trivially identified and pruned before making a prediction.
As a result, if we use \texttt{stx} to denote the event that the sampled program
is syntactically correct, what we care about modeling correctly is
$p( \lambda \ | \ioset{k}{i}{1..K}, \mathtt{stx} )$.
Using Bayes rule, we can rewrite this:
\begin{equation}
  \label{eq:bayes}
  \begin{split}
  p\left( \ \lambda \ |\ \ioset{k}{i}{1..K}, \mathtt{stx}\right) &\propto p\left(\ \mathtt{stx}\ |\ \ioset{k}{i}{1..K}, \lambda\ \right) \times p\left(\ \lambda \ |\ \ioset{k}{i}{1..K}\ \right)\\
  &\propto p\left(\ \mathtt{stx}\ |\ \lambda\ \right) \times p\left(\ \lambda \ |\  \ioset{k}{i}{1..K}\right)
  \end{split}
\end{equation}
We drop the conditional dependency on the IO pairs in the second line
of~(\ref{eq:bayes}) as the syntactical correctness of the program is independent
from the specification when conditioned on the program. We can do the same operation at the token level, denoting by
$\mathtt{stx}_{1 \dotsc t}$ the event that the sequence of the first $t$ tokens $s_1 \dotsb s_t$ doesn't
contain any syntax error and may therefore be a prefix to a valid program.
\begin{equation}
  \label{eq:tokenbayes}
  \begin{split}
    p\left( \ s_t \ |\ s_1 \dotsb s_{t-1}, \ioset{k}{i}{1..K}, \mathtt{stx}_{1 \dotsc t}\right) \propto \quad& p\left( \ \mathtt{stx}_{1 \dotsc t}\ | s_1\dotsb s_t\right) \times \\
   & \quad \ p\left(\ s_t\ |\ s_1 \dotsb s_{t-1}, \ioset{k}{i}{1..K}\right)
  \end{split}
\end{equation}
Given a grammar, it is possible to construct a checker to determine
valid prefixes of programs. Example applications include
compilers to signal syntax errors to the user and autocomplete features of
Integrated Development Environments (IDEs) to restrict the list of suggested
completions.

The quantity $ p\left( \ \mathtt{stx}_{1 \dotsc t}\ |\ s_1\dotsb s_t\right)$ is
therefore not a probability but can be implemented as a deterministic process
for a given target programming language. In practice, this is implemented by
getting at each timestep a mask \mbox{$M = \{-\inf, 0\}^{ | \ \Sigma \ |}$}
where $M_j = -\inf$ if the $j$-th token in the alphabet is not a valid token in
the current context, and 0 otherwise. This mask is added to the output of the
network, just before the Softmax operation that normalizes the output to a
probability over the tokens.

Conditioning on the syntactical correctness of the programs provides several
advantages: First, sampled programs become syntactically correct by
construction. At test time, it allows the beam search to only explore useful
candidates. It also ensures that when we are optimizing for correctness, the
samples used to approximate the distribution are all going to be valid
candidates. Restricting the dimension of the space on which our model is defined
also makes the problem simpler to learn.

\subsection{Jointly learned syntax}
\label{subsec:jointGrammar}
It may not always be feasible to assume access to a syntax checker. In general,
we wish to retain the syntax checker's ability to aggressively prune the search
in program space without requiring access to the syntax itself. To this end, we
propose to represent the syntax checker as a neural network module and learn it
jointly. Similar to the base model, we implement learned syntax checking using
an LSTM $g_\phi$. Comparing to the decoder LSTM, there are two major
differences:
\begin{itemize}
\item The syntaxLSTM is conditioned only on the program tokens, not on the IO pairs. This ensures that the learned checker models only the syntax of the language.
\item The output of the syntaxLSTM is passed through an elementwise $x \mapsto -
  \exp(x)$ activation function and added to the decoder LSTM's output. Similar
  to the mask in Section~\ref{subsec:handwritten}, the exponential activation
  function allows the syntaxLSTM to output high penalties to any tokens deemed
  syntactically incorrect.
\end{itemize}

The addition of the syntaxLSTM doesn't necessitate any change to the training
procedure as it is simply equivalent to a change of architecture. However, in
the supervised setting, when we have access to syntactically correct programs,
we have the possibility of adding an additional term to the loss
~(\ref{eq:mle-obj}) to prevent the model from masking valid programs:
\begin{equation}
  \label{eq:stx_loss}
  \mathcal{L}_{\mathtt{syntax}} = - \sum_{i=1 \dotsb N} \sum_{t= 1 \dotsb L} g_{\phi}\left(\ s^i_t\  | s^i_1 \dotsb s^i_{t-1}\right), \qquad \text{ where } \lambda_i = [s^i_1 , s^i_2 \dotsb s^i_L]
\end{equation}
This loss penalizes the syntaxLSTM for giving negative scores to each token
belonging to a known valid program. We use the reference programs as example of
valid programs when we perform supervised training.

\section{Experiments}
\subsection{The domain: Karel}
\label{subsec:exp_domain}
The Karel programming language is an educational programming
language~\citep{pattis1981karel}, used for example in Stanford CS introductory
classes~\citep{cs106} or in the Hour of Code initiative~\citep{hoc}. It features
an agent inside a gridworld (See Figure \ref{fig:progsynth}), capable of moving
(\texttt{move}, \texttt{turn\{Left,Right\}}), modifying world state
(\texttt{\{pick,put\}Marker}), and querying the state of the nearby environment
for its own markers (\texttt{markerPresent}, \texttt{noMarkerPresent}) or for
natural obstacles (\texttt{frontIsClear}, \texttt{leftIsClear},
\texttt{rightIsClear}).
  Our goal is to learn to generate a program in the Karel
DSL given a small set of input and output grids. The language supports for
loops, while loops, and conditionals, but no variable assignment. Compared to
the original Karel language, we only removed the possibility of defining
subroutines. The specification for the DSL can be found in
appendix~\ref{sec:karel-spec}.

To evaluate our method, we follow the standard practice~\citep{RobustFill,
  NeurosymbProgsynth,neelakantan2015neural, balog2016deepcoder} and use a
synthetic dataset generated by randomly sampling programs from the DSL. We
perform a few simple heuristic checks to ensure generated programs have
observable effect on the world and prune out programs performing spurious
actions (e.g. executing a \texttt{turnLeft} just after a \texttt{turnRight} for
example). For each program, we a set of IO pairs are generated by sampling
random input grids and executing the program on them to obtain the corresponding
output grids. A large number of them are sampled and 6 are kept for each
program, ensuring that all conditionals in a program are hit by at least one of
the examples. The first 5 samples serve as the specification, and the sixth one
is kept as held-out test pair. 5000 programs are not used for training, and get
split out between a validation set and a test set.

We represent the input and output elements as grids where each cell in the grid
is a vector with 16 channels, indicating the presence of elements
(\texttt{AgentFacingNorth}, \texttt{AgentFacingSouth}, $\cdots$,
\texttt{Obstacle}, \texttt{OneMarkerPresent}, \texttt{TwoMarkersPresent},
$\cdots$). The input and output grids are initially passed through independent
convolution layers, before being concatenated and passed through two
convolutional residual blocks and a fully connected layer, mapping them to a
final 512-dimensional representation. We run one decoder per IO pair and perform
a maxpooling operation over the output of all the decoders, out of which we
perform the prediction of the next token. Our models are implemented using the
Pytorch framework~\citep{pytorch}. Code and data will be made available.

\subsection{Results}
We trained a variety of models on the full Karel dataset containing 1-million
examples as well as a reduced dataset containing only $10,000$ examples. In
general, the small dataset serves to help understand the data efficiency of the
program synthesis methods and is motivated by the expected difficulty of
obtaining many labeled examples in real-world program synthesis domains.

The Karel DSL was previously used by ~\citet{metainduction} to study the
relative perfomances of a range of methods depending on the available amount of
data. The task considered was however different as they attempted to perform
program induction as opposed to program synthesis. Rather than predicting a
program implementing the desired transformation, they simply output a
specification of the changes that would result from applying the program so a
direct number comparison wouldn't be meaningful.

Models are grouped according to training objectives. As a baseline, we use
\textbf{MLE}, which corresponds to the maximum likelihood objective
(Eq.\ref{eq:mle-obj}), similar to the method proposed by ~\citet{RobustFill}.
Unless otherwise specified, the reward considered for our other methods is
generalization: +1 if the program matches all samples, including the held out one
and 0 otherwise. \textbf{RL} uses the expected reward objective
(Eq.\ref{eq:RL}), using REINFORCE to obtain a gradient
estimate  (Eq.\ref{eq:Rl-sample}). \textbf{RL\_beam} attempts to solve the proxy
problem described by Equation~\eqref{eq:rl-bs} and \textbf{RL\_beam\_div} the
richer loss function of Equation~\eqref{eq:rl-div-gen}.
\textbf{RL\_beam\_div\_opt} also optimizes the loss of
equation~\eqref{eq:rl-div-gen} but the reward additionally includes a term
inversly proportional to the number of timesteps it takes for the program to
finish executing. All RL models are initialized from pretrained supervised
models.

\setlength{\floatsep}{5pt}
\begin{table}
  \small
  \renewcommand{\arraystretch}{1}
  \setlength{\tabcolsep}{0.7ex}
  \begin{tabular}{ l@{\hspace*{3ex}}cc@{\hspace*{3ex}}cc}
    \toprule
    & \multicolumn{2}{c}{\textbf{Full Dataset}} & \multicolumn{2}{c}{\textbf{Small Dataset}} \\
    \cmidrule(lr{3ex}){2-3}\cmidrule(lr{3ex}){4-5}
    \textbf{Top-1} & \multicolumn{1}{l}{\textbf{Generalization}} & \multicolumn{1}{l}{\textbf{Exact Match}} & \multicolumn{1}{l}{\textbf{Generalization}} & \multicolumn{1}{l}{\textbf{Exact Match}} \\
    \midrule
    \textbf{MLE} & 71.91 & \textbf{39.94} & 12.58 & 8.93 \\
    \textbf{RL} & 68.39 & 34.74 & 0 & 0 \\
    \textbf{RL\_beam} & 75.72 & 8.21 & \textbf{25.28} & \textbf{17.63} \\
    \textbf{RL\_beam\_div} & 76.20 & 31.25 & 23.72 & 16.31 \\
    \textbf{RL\_beam\_div\_opt} & \textbf{77.12} & 32.17 & 24.24 & 16.63 \\
    \bottomrule
  \end{tabular}
  \caption{\textbf{RL\_beam} optimization of program correctness results in
    consistent improvements in top-1 generalization accuracy over supervised
    learning \textbf{MLE}, even though the exact match of recovering the
    reference program drops. The improved objective function results in further
    improvements.}
  \label{tab:rl_results}
\end{table}

\paragraph{Optimizing for correctness (RL):\ }\mbox{}
Results in Table \ref{tab:rl_results} show that optimizing for the expected
program correctness consistently provides improvements in top-1 generalization
accuracy. \textbf{Top-1 Generalization Accuracy} (Eq. \ref{eq:generalization})
denotes the accuracy of the most likely program synthesized by beam search
decoding having the correct behaviour across all input-output examples. We
didn't perform any pruning of programs that were incorrect on the 5
specification examples. The improved performance of RL methods confirms our
hypothesis that better loss functions can effectively combat the program aliasing problem.

On the full dataset, when optimizing for correctness, \textbf{Exact Match
  Accuracy} decreases, indicating that the RL models no longer prioritize
generating programs that exactly match the references. On the small dataset,
\textbf{RL\_beam} methods improves both exact match accuracy and generalization.

Comparing RL\_beam to standard RL, we note improvements across all levels of
generalization. By better aligning the RL objective with the sampling that
happens during beam search decoding, consistent improvements can be made in
accuracy. Further improvements are made by encouraging diversity in the beam of
solutions (RL\_beam\_div) and penalizing long running programs
(RL\_beam\_div\_opt).

In the settings where little training data is available, RL methods
show more dramatic improvements over MLE, indicating that data
efficiency of program synthesis methods can be greatly improved by
using a small number of samples first for supervised training and
again for Reinforcement Learning.

As a side note, we were unable to achieve high performance when
training the RL methods from scratch. The necessity of extensive
supervised pretraining to get benefits from Reinforcement Learning
fine-tuning is well-known in the Neural Machine Translation
literature~\citep{ranzato2015sequence,googleNMT,
  wiseman2016sequence,bahdanau2016actor}.

Table \ref{tab:beam_results} examines the top-1, top-5, and top-50
generalization accuracy of supervised and RL models. RL\_beam methods performs
best for top-1 but their advantage drops for higher-rank accuracy. Inspection of
generated programs shows that the top predictions all become diverse
variations of the same program, up to addition/removal of no-operations
(\texttt{turnLeft} followed by \texttt{turnRight}, full circle obtained by a
series of four \texttt{turnLeft}). The RL\_beam\_div objective helps alleviate
this effect as does RL\_beam\_div\_opt, which penalizes redundant programs.
This is important as in the task of Program Synthesis, we may not necessarily
need to return the most likely output if it can be pruned by our specification.

\begin{table}
  \begin{floatrow}
    \capbtabbox[.4\textwidth]{%
      \small
      \renewcommand{\arraystretch}{1}
      \setlength{\tabcolsep}{0.7ex}
      \begin{tabular}{ l@{\hspace*{1ex}}c@{\hspace*{1ex}}c@{\hspace*{1ex}}c}
        \toprule
        \textbf{Generalization} & Top-1 & Top-5 & Top-50 \\
        \midrule
        \textbf{MLE} & 71.91 & 79.56 & \textbf{86.37} \\
        \textbf{RL\_beam} & 75.72 & 79.29 & 83.49 \\
        \textbf{RL\_beam\_div} & 76.20 & 82.09 & 85.86 \\
        \textbf{RL\_beam\_div\_opt} & \textbf{77.12} & \textbf{82.17} & 85.38 \\
        \bottomrule
      \end{tabular}
    }{%
      \caption{\label{tab:beam_results}\textbf{Top-k accuracies: } MLE shows
        greater relative accuracy increases as k increases than RL. Methods
        employing beam search and diversity objectives reduce this accuracy
        gap by encouraging diversity in the beam of partial programs.}
    }
    \hfill
    \capbtabbox[.55\textwidth]{%
      \small
      \renewcommand{\arraystretch}{1}
      \setlength{\tabcolsep}{0.7ex}
      \begin{tabular}{ l@{\hspace*{3ex}}ccc@{\hspace*{3ex}}ccc  @{\hspace*{6ex}}ccc @{\hspace*{3ex}}ccc}
        \toprule
        \textbf{Top-1 Generalization} & \textbf{Full Dataset} & \textbf{Small Dataset} \\
        \midrule
        \textbf{MLE} & 71.91 & 12.58 \\
        \textbf{MLE\_learned} & 69.37 & \textbf{17.02} \\
        \textbf{MLE\_handwritten} & 72.07 & 9.81 \\
        \textbf{MLE\_large} & \textbf{73.67} & 13.14 \\
        \bottomrule
      \end{tabular}
    }{%
      \caption{\label{tab:syntax_results}\textbf{Grammar prunes the space of
          possible programs}: On the full dataset, handwritten syntax checking
        MLE\_handwritten improves accuracy over no grammar MLE, although
        MLE\_large shows that simply adding more parameters results in even
        greater gains. On the small dataset, learning the syntax MLE\_learned
        outperforms handwritten grammar and larger models.}
    }
  \end{floatrow}
\end{table}

\paragraph{Impact of syntax:}\mbox{}
We also compare models according to the use of syntax:
\textbf{MLE\_handwritten} denotes the use of a handwritten syntax
checker (Sec \ref{subsec:handwritten}), \textbf{MLE\_learned} denotes
a learned syntax (Sec \ref{subsec:jointGrammar}), while no suffix
denotes no syntax usage. Table \ref{tab:syntax_results} compares
syntax models.

On the full dataset, leveraging the handwritten syntax leads to
marginally better accuracies than learning the syntax or using no
syntax. Given access to enough data, the network seems to be capable
of learning to model the syntax using the sheer volume of training
examples.

On the other hand, when the amount of training data is limited,
learning the syntax produces significantly better performance. By
incorporating syntactic structure in the model architecture and
objective, more leverage is gained from small training
data. Interestingly, the learned syntax model even outperforms the
handwritten syntax model. We posit the syntaxLSTM is free to learn a
richer syntax. For example, the syntaxLSTM could learn to model the
distribution of programs and discourage the prediction of not only
syntactically incorrect programs, but also the unlikely ones.

To control for the extra parameters introduced by the syntaxLSTM, we
compare against MLE\_large, which uses no syntax but features
a larger decoder LSTM, resulting in the same number of parameters as
MLE\_learned. Results show that the larger number of parameters is not
enough to explain the difference in performance, which again indicates
the utility of jointly learning syntax.

\paragraph{Analysis of learned syntax:}\mbox{}
Section~\ref{subsec:jointGrammar} claimed that by decomposing our
models into two separate decoders, we could decompose the learning so
that one decoder would specialize in picking the likely tokens given
the IO pairs, while the other would enforce the grammar of the
language. We now provide experimental evidence that this decomposition
happens in practice.
\setcounter{figure}{1}
\begin{wrapfigure}[25]{R}{.4\textwidth}
  \floatbox[\nocapbeside]{table}[\FBwidth]
  {\caption{Importance of Syntax}\label{tab:syntaxDecomp}}
  {
    \noindent\resizebox{\linewidth}{!}{
    \begin{tabular}{@{}c@{\hspace*{5ex}}cc@{}}
      \toprule
      \twoLines{\% Synctactically}{Correct}& \twoLines{Joint}{Model} & \twoLines{Without}{Learned Syntax}\\
      \midrule
      Amongst Top1 & 100 \% & 0 \% \\
      Amongst Top5 & 100 \% & 0 \% \\
      Amongst Top50 & 100 \% & 0 \% \\
      Amongst Top100 & 99.79 \% & .04 \% \\
      \bottomrule
    \end{tabular}}
    }
    \vspace{5pt}
    \ffigbox{\caption{Syntax Comparison}\label{fig:grammarComp}}
    {\begin{subfloatrow}
        \ffigbox[\FBwidth][][]{\caption{Manual}\label{subfig:manGrammar}}
        {\includegraphics[width=.3\textwidth]{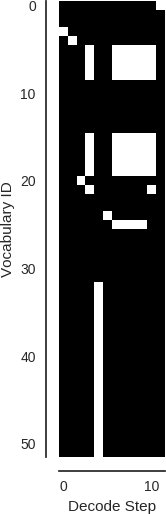}}
        \ffigbox[\FBwidth][][]{\caption{Learned}\label{subfig:learnedGrammar}}
        {\includegraphics[width=.3\textwidth]{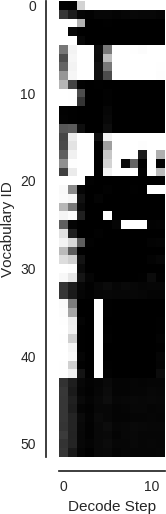}}
        \ffigbox[\FBwidth][][]{\caption{Diff}\label{subfig:diffGrammar}}
        {\includegraphics[width=.3\textwidth]{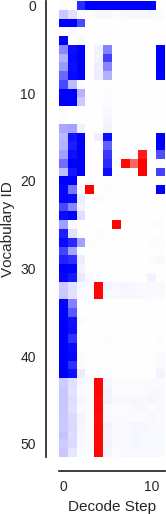}}
      \end{subfloatrow}
      \caption{\label{fig:grammarComp}Syntax Comparison}
    }
  \end{wrapfigure}

Table~\ref{tab:syntaxDecomp} shows the percentage of syntactically
correct programs among the most likely predictions of the \textbf{MLE
  + learned} model trained on the full dataset. Both columns
correspond to the same set of parameters but the second column doesn't
apply the syntaxLSTM's mask to constrain the decoding process. The
precipitous drop in syntax accuracy indicates the extent to which the
program decoder has learned to rely on the syntaxLSTM to produce
syntactically correct programs.

Figure~\ref{fig:grammarComp} compares the syntax masks generated by
the learned and handwritten syntax models while decoding Program A in
Figure~\ref{fig:progsynth}. (\ref{subfig:manGrammar}) shows output of
the handwritten syntax checker; (\ref{subfig:learnedGrammar}) shows
syntaxLSTM output. White cells indicates tokens that are labeled
syntactically correct at each decoding step.

Figure \ref{subfig:diffGrammar} analyzes the difference between the
handwritten and learned syntax masks. White indicates similar output,
which occupies the majority of the visualization. Blue cells
correspond to instances where the syntaxLSTM labeled a token correct
when it actually was syntactically incorrect. This type of error can
be recovered if the program decoder predicts those tokens as unlikely.

On the other hand, red cells indicate the syntaxLSTM predicted a valid
token is syntactically incorrect. This type of error is more dangerous
because the program decoder cannot recover the valid token once it is
declared incorrect. The majority of red errors correspond to tokens
which are rarely observed in the training dataset, indicating that the
syntaxLSTM learns to model more than just syntax - it also captures
the distribution over programs. Given a large enough dataset of real
programs, the syntaxLSTM learns to consider non-sensical and unlikely
programs as syntactically incorrect, ensuring that generated programs
are both syntactically correct and likely.

\section{Conclusion}
We presented two novel contributions to improve state-of-the-art neural program
synthesis techniques. Our first contribution uses Reinforcement Learning to
optimize for generating any consistent program, which we show helps in improving
generalization accuracy of the learned programs for large training datasets. Our
second contribution incorporates syntax checking as an additional conditioning
mechanism for pruning the space of programs during decoding. We show that
incorporating syntax leads to significant improvements with limited training
datasets.

\clearpage
{\small
\bibliographystyle{plainnat}
\bibliography{bibliography}
}
\clearpage
\appendix
\section{Computing the richer loss function}
\label{app:div_details}
In this section, we describe how the objective function of
Equation~\eqref{eq:rl-div-gen} can be computed.

We have a distribution $q_{\theta}$ over $S$ programs, as obtained by performing
a beam search over $p_{\theta}$ and renormalizing. We are going to get $C$
independent samples from this distribution and obtain a reward corresponding to
the best performing of all of them.

In the case where the reward function $R_i(\lambda_r)$ is built on a boolean
property, such as for example ``correctness of the generated program'',
Equation~\eqref{eq:rl-div-gen} can be simplified. As $R_i(\lambda_r)$ can only
take the values 0 or 1, the term $\max_{j \in 1..C} R_i(\lambda_j)$ is going to
be equal to 0 only if all of the $C$ sampled programs give a reward of zero. For
each sample, there is a probability of
\mbox{$q_{\text{incorrect}}=\sum_{\lambda_r \in \text{BS}(p_{\theta}, S)}
  [R_i(\lambda_r) == 0]\ q_{\theta}(\lambda_r \ | \ioset{k}{i}{1..K})$} of
sampling a program with rewards zero. The probability of not sampling a single
correct program out of the $C$ samples is $q_{\text{incorrect}}^C$. From this,
we can derive the form of Equation~\eqref{eq:rl-div-zeroone}. Note that this can
be computed without any additional sampling steps as we have a close form
solution for this expectation.

In the general case, a similar derivation can be obtained. Assume that the
programs outputted by the beam search have associated rewards $R_0, R_1, ...,
R_S$ and assume without loss of generality that $R_0 < R_1 < .. < R_S$. The
probability of sampling a program with a reward smaller than $R_i$ is
\mbox{$q_{\leq R_i}=\sum_{\lambda_r \in \text{BS}(p_{\theta}, S)}
  [R_i(\lambda_r) \leq R_i]\ q_{\theta}(\lambda_r \ | \ioset{k}{i}{1..K})$} so
the probability of obtaining a final reward of less than $R_i$ when sampling C
samples is $q_{\leq R_i}^C$. As a result, the probability of obtaining a reward
of exactly $R_i$ is $\left( q_{\leq R_i}^C - q_{\leq R_{i-1}}^C \right)$. Based
on this, it is easy to evaluate (and differentiate) the loss function described by
Equation~\eqref{eq:rl-div-gen}.

\FloatBarrier
\section{Karel Langage specification}
\label{sec:karel-spec}
\begin{figure}[htp]
  \begin{center}
    \begin{eqnarray*}
      \mbox{Prog } p & := & \mathtt{def }\ \mathtt{run}()\;:\ s \\
      \mbox{Stmt } s & := & \mathtt{while}(b): s \; |\  \mathtt{repeat}(r): s \;
                            |\ s_1; s_2 \; | \; a \\
                     & | & \mathtt{if}(b): s \; | \; \mathtt{ifelse}(b): s_1\ \mathtt{else}: s_2 \\
      \mbox{Cond } b & := & \mathtt{frontIsClear()} \; | \; \mathtt{leftIsClear()} \; | \; \mathtt{rightIsClear()} \\
                     & | & \mathtt{markersPresent()} \; | \;
                           \mathtt{noMarkersPresent()} \ | \ \; \mathtt{not} \; b  \\
      \mbox{Action a} & := & \mathtt{move()} \; | \; \mathtt{turnRight()} \; | \; \mathtt{turnLeft()} \\
                     & | & \mathtt{pickMarker()} \; | \; \mathtt{putMarker()} \\
      \mbox{Cste } r & := & 0 \; |\ 1 \; | \ \dotsb \; |\  19
    \end{eqnarray*}
  \end{center}
  \caption{The Domain-specific language for Karel programs.}
  \label{fig:karelDSL}
\end{figure}

\FloatBarrier
\section{Experiments hyperparameters}
The state of the grid word are represented as a $16 \times 18 \times 18$ tensor.
For each cell of the grid, the 16-dimensional vector corresponds to the feature
indicated in Table~\ref{fig:grid-rep}
\begin{table}[h]
  \begin{tabular}{|c|}
    \hline
    Hero facing North \\
    \hline
    Hero facing South \\
    \hline
    Hero facing West \\
    \hline
    Hero facing East \\
    \hline
    Obstacle \\
    \hline
    Grid boundary \\
    \hline
    1 marker \\
    \hline
    2 marker \\
    \hline
    3 marker \\
    \hline
    4 marker \\
    \hline
    5 marker \\
    \hline
    6 marker \\
    \hline
    7 marker \\
    \hline
    8 marker \\
    \hline
    9 marker \\
    \hline
    10 marker \\
    \hline
  \end{tabular}
  \caption{Representation of the grid}
  \label{fig:grid-rep}
\end{table}

\begin{table}[h]
  \centering
  \begin{tabular}{|c||c|c|}
    \hline
    & \textbf{Input Grid} & \textbf{Output Grid}\\
    \cline{2-3}
    \multirow{2}{*}{Grid Embedding} & Conv2D, kernel size = 3, padding = 1, 16 $\rightarrow$ 32 & Conv2D, kernel size = 3, padding = 1, 16 -> 32\\
    & ReLU & ReLU \\
    \hline
    \multirow{6}{*}{Residual Block 1} & \multicolumn{2}{c|}{Conv2D, kernel size = 3, padding 1, 64 $\rightarrow$ 64}\\
 & \multicolumn{2}{c|}{ReLU}\\
 & \multicolumn{2}{c|}{Conv2D, kernel size = 3, padding 1, 64 $\rightarrow$ 64}\\
 & \multicolumn{2}{c|}{ReLU}\\
 & \multicolumn{2}{c|}{Conv2D, kernel size = 3, padding 1, 64 $\rightarrow$ 64}\\
 & \multicolumn{2}{c|}{ReLU}\\
    \hline
    \multirow{6}{*}{Residual Block 1} & \multicolumn{2}{c|}{Conv2D, kernel size = 3, padding 1, 64 $\rightarrow$ 64}\\
 & \multicolumn{2}{c|}{ReLU}\\
 & \multicolumn{2}{c|}{Conv2D, kernel size = 3, padding 1, 64 $\rightarrow$ 64}\\
 & \multicolumn{2}{c|}{ReLU}\\
 & \multicolumn{2}{c|}{Conv2D, kernel size = 3, padding 1, 64 $\rightarrow$ 64}\\
 & \multicolumn{2}{c|}{ReLU}\\
    \hline
    Fully Connected & \multicolumn{2}{c|}{Linear, 20736 $\rightarrow$ 512}\\
    \hline

  \end{tabular}
  \caption{Encoding of the Input/Output Pairs}
\end{table}

The decoders are two-layer LSTM with a hidden size of 256. Tokens of the DSL are
embedded to a 256 dimensional vector. The input of the LSTM is therefore of
dimension 768 (256 dimensional of the token + 512 dimensional of the IO pair
embedding). The LSTM used to model the syntax is similarly sized but doesn't
take the embedding of the IO pairs as input so its input size is only 256.

One decoder LSTM is run on the embedding of each IO pair. The topmost activation
are passed through a MaxPooling operation to obtain a 256 dimensional vector
representing all pairs. This is passed through a linear layer to obtain a score
for each of the 52 possible tokens.
We add the output of the model and of the eventual syntax model, whether learned
or handwritten and pass it through a SoftMax layer to obtain a probability
distribution over the next token.

All training is performed using the Adam optimizer, with a learning rate of
$10^-4$. Supervised training used a batch size of 128 and RL methods used a
batch size of 16. We used 100 rollouts per samples for the Reinforce method and
a beam size of 64 for methods based on the beam search. The value of C used for
the methods computing a loss on bags of programs was 5.

\end{document}